\documentclass[11pt]{article}
\usepackage[parfill]{parskip}
\usepackage{longtable}
\usepackage[dvipsnames]{xcolor}
\usepackage{booktabs}

\newcommand{\dssat}{\textsc{dssat}}
\newcommand{\swap}{\textsc{swap}}

\providecommand{\tightlist}{%
	  \setlength{\itemsep}{0pt}\setlength{\parskip}{0pt}}

\usepackage{graphicx}
\usepackage{xspace}
\usepackage[top=1in, bottom=1in, left=1in, right=1in]{geometry}

\usepackage{amsmath, amssymb}

\usepackage{hyperref}
\hypersetup{
  unicode=true,
  linktocpage=true,
  linkcolor=NavyBlue,
  colorlinks=true,
  urlcolor=NavyBlue,
  citecolor=RoyalBlue,
  breaklinks=true,
}

\usepackage{fancyhdr}
\usepackage{titlesec}
\titlespacing{\section}{0pt}{\parskip}{-.5\parskip}
\titlespacing{\subsection}{0pt}{\parskip}{- .5\parskip}
\titlespacing{\subsubsection}{0pt}{\parskip}{- .5\parskip}

\date{\vspace{-5ex}}

\usepackage{authblk}
\author[1]{Adarsh Pyarelal}
\author[2]{Marco A. Valenzuela-Esc\'arcega}
\author[2]{Rebecca Sharp}
\author[2]{Paul D. Hein}
\author[2]{Jon Stephens}
\author[2]{Pratik Bhandari}
\author[2]{HeuiChan Lim}
\author[2]{Saumya Debray}
\author[1]{Clayton T. Morrison}
\affil[1]{School of Information, University of Arizona, Tucson, AZ}
\affil[2]{Department of Computer Science, University of Arizona, Tucson, AZ}

\title{AutoMATES: Automated Model Assembly from Text, Equations, and Software}

\begin{document}
\maketitle
\vspace{10pt}
\begin{center}
\href{https://ml4ai.github.io/automates}{\texttt{ml4ai.github.io/automates}}
\end{center}
\begin{abstract} 

  Models of complicated systems can be represented in different ways - in
  scientific papers, they are represented using natural language text as well as
  equations. But to be of real use, they must also be implemented as software, thus
  making code a third form of representing models.
  We introduce the AutoMATES project, which aims to build semantically-rich
  unified representations of models from scientific code and publications to 
  facilitate the integration of computational models from different domains and
  allow for modeling large, complicated systems that span multiple domains and
  levels of abstraction.


\end{abstract}

\section{Introduction}
\label{sec:intro}

There exist today state-of-the-art computational models that can provide highly
accurate predictions about complex phenomena such as crop growth and weather
patterns. However, certain phenomena, such as food insecurity, involve a host of
factors that cannot be modeled by any single one of these models, but which
instead require the integration of multiple models.

To truly integrate these computational models, it is necessary to `lift' them to
a common representation that is (i) agnostic to the software implementation,
(ii) semantically rich enough to represent the implicit domain knowledge in the
models, and (iii) connected to the domain literature.
The AutoMATES project aims to build technology to construct and curate
semantically-rich representations of scientific models by integrating three
different sources of information:

\begin{itemize}
  \tightlist
  \item natural language descriptions of models in publications and other technical documentation,
  \item the equations contained in these documents, and
  \item the software the implements these models.
\end{itemize}

\begin{figure}[t]
  \centering
  \includegraphics[width=\textwidth]{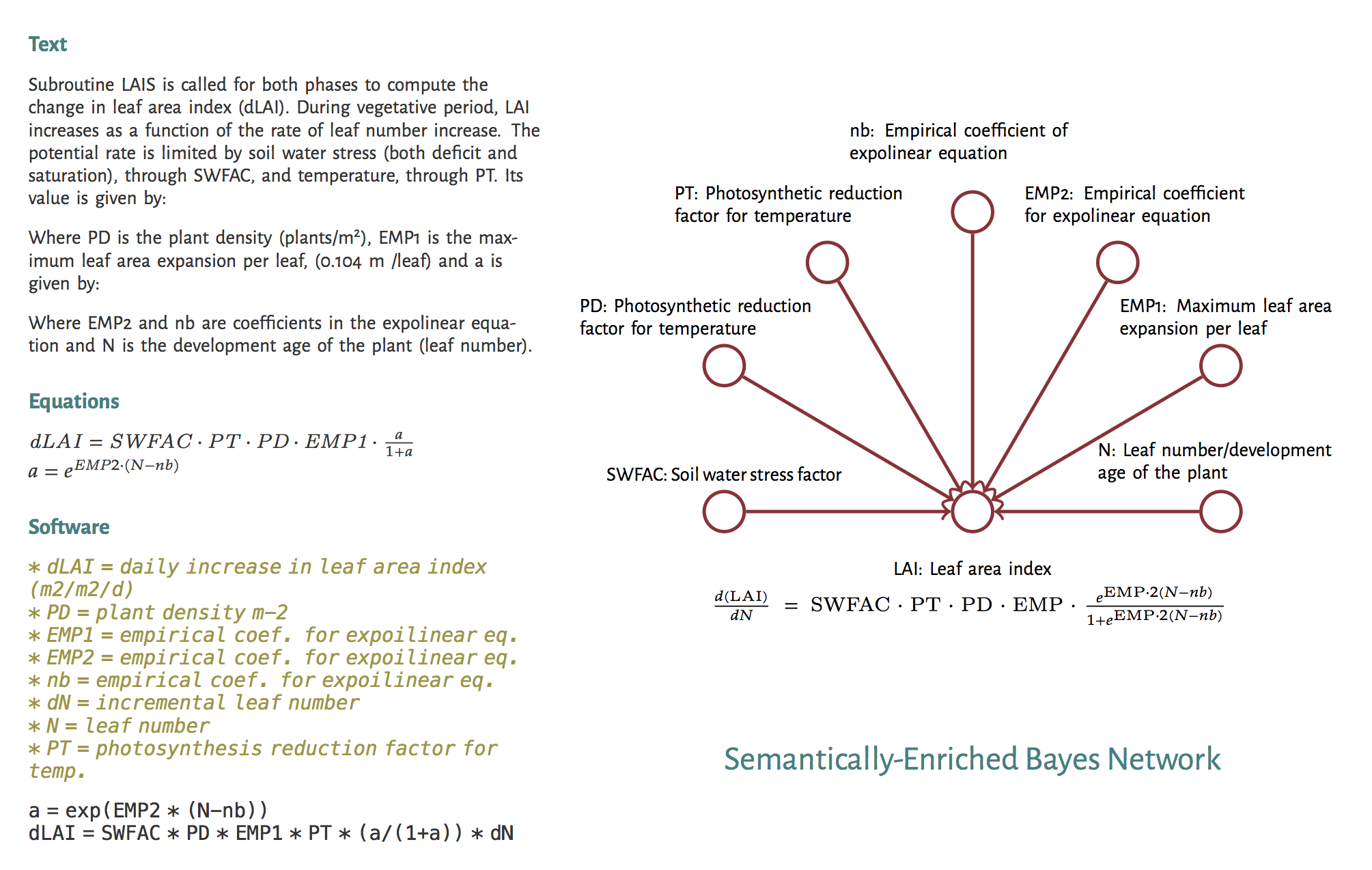}
  \caption{Integration of text, equations, and code into a semantically-enriched Bayesian network.}
  \label{fig:confluence_example}
\end{figure}

An example of a model being represented in these three forms (text, equations,
and software) is shown in \autoref{fig:confluence_example}. This model is a
differential equation describing the biophysical variable, leaf area index
(LAI). The network on the right half of the figure is an aspirational
representation of the model as a Bayesian network. Although this example is
hand-crafted, our end goal is to be able to automatically assemble models with
this level of semantic richness.
In this paper, we describe our high-level approach%
\footnote{For more technical details, please visit
  \href{https://ml4ai.github.io/automates/documentation/deliverable_reports}{\texttt{ml4ai.github.io/automates/documentation/deliverable\_reports}.}} and present our latest results. 

\paragraph{Significance:}
This work will dramatically advance the state-of-the-art in automated model
curation and integration, enabling scientists and analysts to understand complex
mechanisms that span multiple domains. By exposing the implicit domain knowledge
baked into computational models, this effort will enable semantically rich automated model
composition and reasoning in context, at scale.

\hypertarget{architecture-overview}{%
\section{Architecture Overview}\label{architecture-overview}}

\begin{figure}[t]
  \centering
  \includegraphics[width=\textwidth]{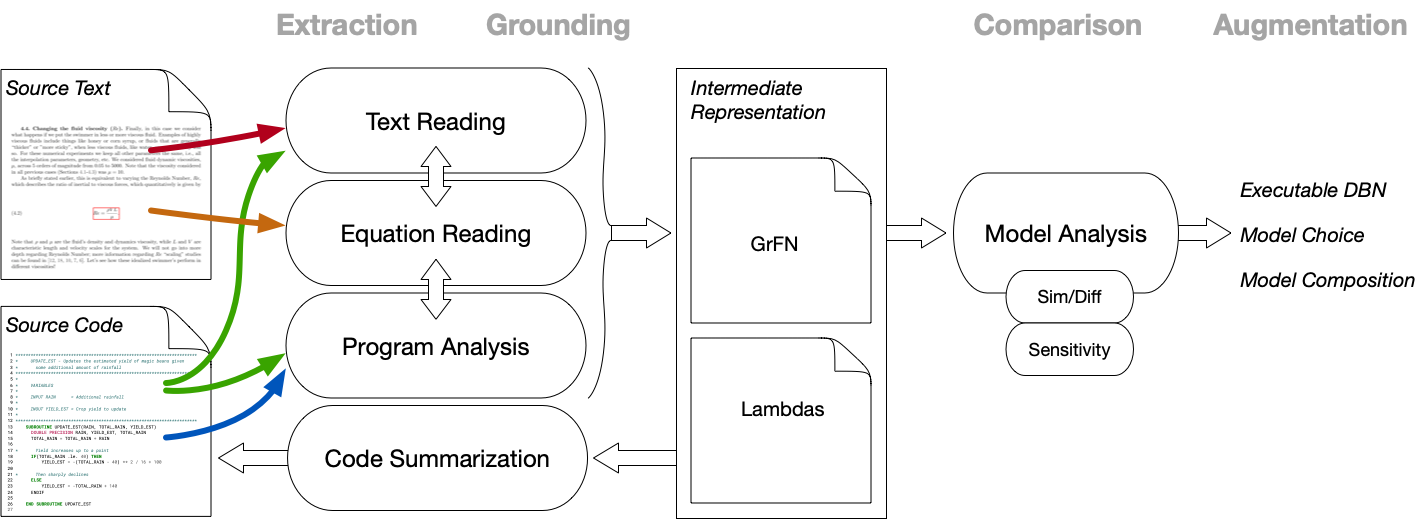}
  \caption{Architecture overview}
    \label{fig:arch}
\end{figure}

The AutoMATES system is designed to extract information from several knowledge
sources, link the extracted concepts into a unified model representation,
compare models based on different features, and augment the models with
supplementary information.  Each of these components is implemented
independently, but designed to interoperate with each other.  A high-level view
of the system's architecture is shown in \autoref{fig:arch}.  This modular
architecture provides AutoMATES with the extensibility required to support
different knowledge sources in the future as it is extended to handle other
domains. Briefly, the four main components of AutoMATES are:

\begin{enumerate}
\def\labelenumi{\arabic{enumi}.}
\tightlist

\item \emph{Extracting} model information from different aspects of source code
  and corresponding scientific publications and technical documents.  From source code, we extract
  the model from the code implementation itself
  (\autoref{sec:program_analysis}) as well as any supplementary information
  such as descriptions expressed in comments.  From scientific publications and documentation,
  AutoMATES reads model information from text (\autoref{sec:reading}) and
  equations (\autoref{sec:equations}).

\item \emph{Grounding} the extracted information by identifying when the
  same concept is expressed in different knowledge sources, and linking them
  together to form a unified, programming language-agnostic intermediary model
  representation: a \emph{Grounded Function Network
  (GrFN)}.  
  
\item \emph{Comparing} models, using the GrFN representation,  by
  analyzing structural and functional (via sensitivity analysis) similarities
  and differences (\autoref{sec:model_analysis}). 

\item \emph{Augmenting} models through selection of model components
  appropriate for a task, composing model components, generating model descriptions in context to augment existing documentation, and model execution.
  
\end{enumerate}

\section{Scalable Program Analysis}
\label{sec:program_analysis}

Our program analysis approach to extracting model information from the
source code implementation begins with \texttt{for2py}, a front-end
translator that maps Fortran source programs to a language-independent
program analysis intermediate representation (PAIR) that is then used to
generate files used as input to subsequent analysis.  This design
decouples input processing from output generation, and is motivated by
the following:

\begin{enumerate}
\tightlist
\item \textit{Performance and scalability.} Modules that are referenced
  by multiple program components do not have to be reanalyzed separately
  for each referencing component. Independent source-language modules
  can in principle, be analyzed concurrently.

\item \textit{Support for source-language heterogeneity.} This design
  makes it possible, in principle, to support programs with different
  components written in different languages. It also allows us to reason
  about models implemented in different source languages.

\item \textit{Independence of back-end tasks.} Different back-end
  analysis tasks, e.g., sensitivity analysis and comment analysis, can
  be carried out independently (and, if necessary, concurrently) on the
  PAIR.

\end{enumerate}

\texttt{for2py} currently handles a significant subset of Fortran,
including: data types such as scalars and arrays; control constructs
such as conditionals, loops, functions, and subroutines; and
input/output (I/O) primitives including formatted and list-directed I/O.
We expect to soon complete the handling of modules and derived types.

A fundamental challenge we have to address is that of scalability, since software implementing sophisticated scientific models can encompass thousands of source files and hundreds of thousands of lines of code.  We do this by performing analysis at the module level of granularity.  Given the source code for a scientific model, we analyze its modules to identify define-use relationships between them and construct a \textit{module dependency graph} that identifies these dependencies between different modules.  We use a topological sort of this graph to guide the subsequent analysis of the modules. The module dependency graph imposes a partial order on the modules of the analyzed system, indicating which modules are independent of each other and can therefore be analyzed in parallel. This ordering has three significant implications for scaling. First, it allows modern computer systems such as multi-core processors and cloud-based systems to be utilized effectively. Second, it provides the user a straightforward tunable tradeoff between computational resources and analysis efficiency. Finally, it means that the cost of analyzing a software system is proportional to the depth of its module dependency graph rather than its total size (number of nodes), resulting in sublinear asymptotic complexity.

\section{Contextualized Equation Parsing}
\label{sec:equations}

Models are often represented concisely as equations, at a level of
abstraction that can supplement both the natural language description as
well as the source code implementation.  For humans to compare the
equations and source code for several models, as is done in
\cite{Camargo:2016}, is time consuming and expensive.
Accordingly, we are developing an automated approach that identifies the relevant
equations in text and rendered images of documents (PDFs treated as images) associated with scientific models, parse them into an intermediate symbolic mathematical representation, and ground the variables in the equations to text descriptions and source code variables.  

Non-textual elements in PDFs have previously been identified using heuristics
based on document structure \cite{pdffigures2} or statistical learning
\cite{chu2013mathematical,bruce2014mathematical}.
Here, taking advantage of advances in deep learning \cite{redmon2016you}, we
identify the location of
the bounding box surrounding the equations using machine vision techniques.
After identifying the location of the equations in the PDF, the next step in the
pipeline is to parse the rendered equation into an intermediate representation.
We choose to use \LaTeX{} because we have the \LaTeX{} source code for each of
the training examples, and also because \LaTeX{} preserves all of the typographic
information (e.g., boldface, subscript, etc.), which conveys variable semantics. 
We decompile the image using an encoder-decoder system that
encodes the image of the equation through a series of convolutions and produces
\LaTeX~commands that generate the image~\cite{Deng:2016, Deng:2017}.  
We are currently evaluating this process on a held-out subset of the data from ArXiv.

This decoded \LaTeX{} representation will then be parsed into Python
code (by extending coverage of an open source rule-based system,
\texttt{latex2sympy}\footnote{\url{https://github.com/augustt198/latex2sympy}},
to equation elements frequently found in the domain) and then converted
to a 
equation elements frequently found in the domain) and then converted to a GrFN
representation.
To ground the GrFN representation extracted from the equations, we locate text that
references the equation, using the equation identifier when available and the
lexical content when it is not. 

\begin{figure}
  \centering
  \includegraphics[width=0.7\textwidth]{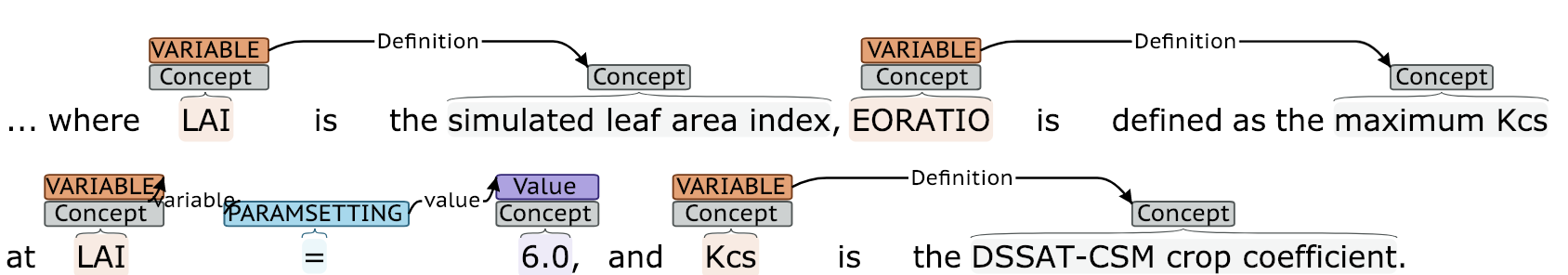}
  \caption{Example of variables (represented as concepts, definitions and value assignments) extracted from scientific text as a result of the machine reading pipeline.
  \label{fig:extractions}}
\end{figure}

\section{Machine Reading for Scientific Models}
\label{sec:reading}

We are developing a framework for reading and extracting model information from the scientific papers that directly describe the computational models (e.g., \dssat{}, \swap{}) whose source code we analyze
(\autoref{sec:program_analysis}).

Scientific papers are typically available as PDFs, which need to be preprocessed into a format that a machine reader can use. We make use of Science Parse\footnote{\url{https://github.com/allenai/science-parse}}, an open source tool that segments the sections based on the paper layout and typography.  

Our framework then implements an open-domain information extraction
system based on Eidos\footnote{\url{https://github.com/clulab/eidos}}, a
machine reading system designed to extract causal relations.  At its core, Eidos
has a grammar of rules~\cite{valenzuela2016odin, Escarcega:2018} that model
linguistic patterns commonly used by authors to express causality in text.
Here, where we are interested in gathering context about the models implemented
in source code, causal relations are useful, but not sufficient.  We have
modified Eidos to extract mentions of model variables and their descriptions.
Additionally, it will be critical to read for background assumptions (e.g.,
model preconditions) and additional contextual information which could inform
the setting of parameters (using quantities and units identified by grobid-quantities\footnote{\url{https://github.com/kermitt2/grobid-quantities}}).

\begin{figure}[t]
  \centering
  \includegraphics[width=0.9\textwidth]{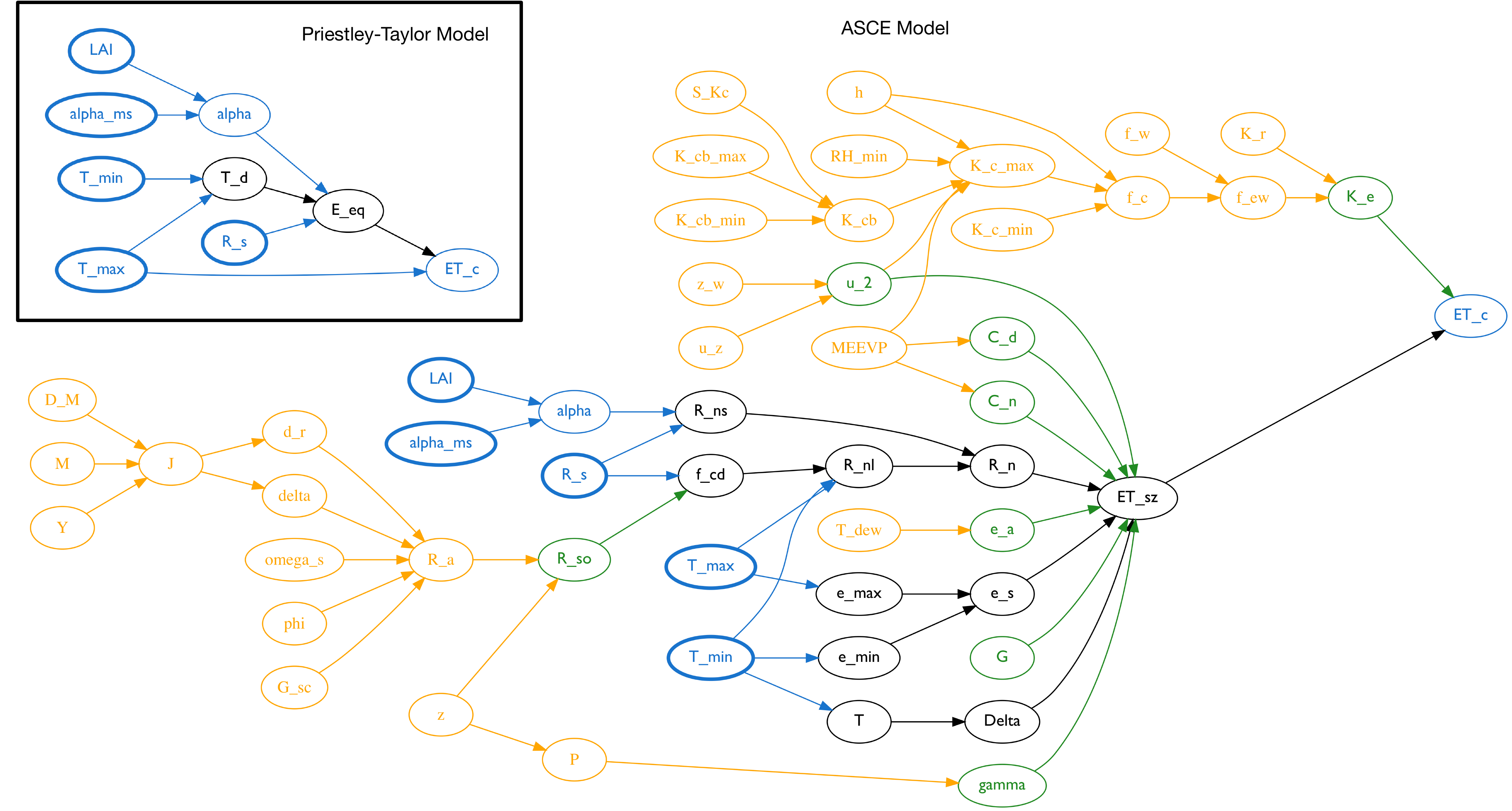}
  \caption{Results of comparing Priestly-Taylor (PT) and ASCE models. Blue nodes represent variables shared between PT and ASCE. Black nodes represent variables not shared but along directed paths between shared variables. Green nodes in the ASCE model represent variables whose states directly affect shared directed paths -- if controlled, this isolates the portions of ASCE that overlap with PT. Finally, orange nodes represent variables in the ASCE model that can be isolated from the overlap in the comparison.
  \label{fig:comparison}}
\end{figure}

The extracted variables and their mentions will necessarily be aligned with
the variables read from source code (\autoref{sec:program_analysis}) 
and equations (\autoref{sec:equations}) to find
and resolve commonalities and discrepancies in different representations of the
same model.
In \autoref{fig:extractions}, we show a screenshot showing results of the
current text reading pipeline.

\begin{figure}[t]
  \centering
  \includegraphics[width=\textwidth]{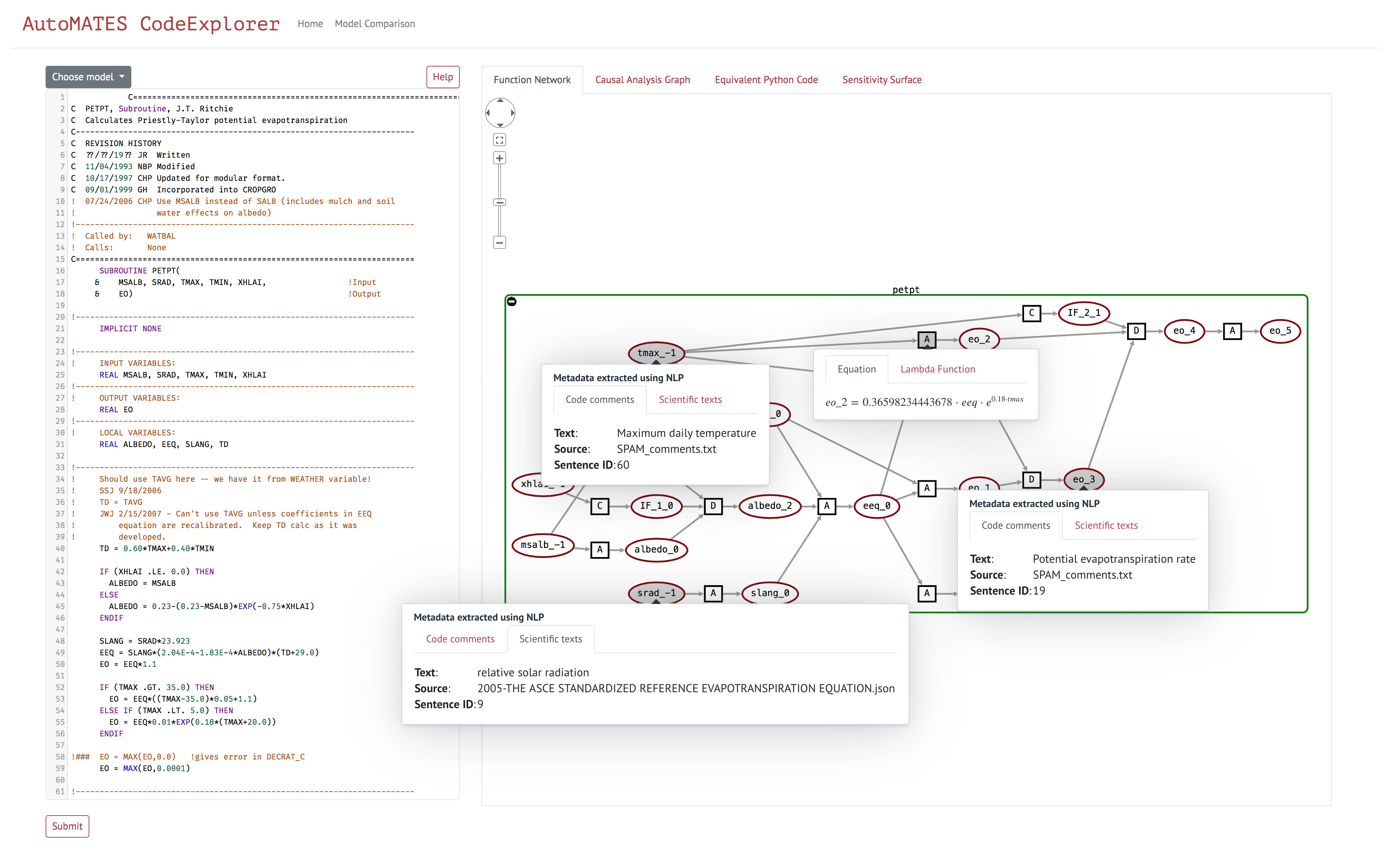}
  \caption{Screenshot of the AutoMATES CodeExplorer (available at
  \url{http://vanga.sista.arizona.edu/automates}), showing the translation of a
  the Priestley-Taylor method for calculating potential evapotranspiration (a
  submodule in DSSAT \cite{DSSAT}) into a computation graph. The
  \texttt{\_\_assign\_\_} nodes are annotated with the automatically extracted
  \LaTeX{}-typeset representation of the equation extracted from the code, which
will facilitate linking with scientific publications. Additionally, the
variable nodes are automatically aligned with descriptions extracted from code
comments and scientific texts.}
    \label{fig:demowebapp}
\end{figure}

\section{Model Analysis}
\label{sec:model_analysis}

Model comparison and eventual augmentation is then enabled by our model
analysis pipeline, which identifies which portions of two or
more models share the same or similar computations about similar variables,
and which components are different.

This analysis is enabled by the unified grounded function network (GrFN)
representation, such that we first identify shared variables and then analyze
the GrFN topology to identify differences in setting variables states.
\autoref{fig:comparison} shows an example of comparing the PT and ASCE
evapotranspiration models from the DSSAT crop modeling system \cite{DSSAT}.
Sensitivity analysis is then used to analyze the functional relationships
between the variables.  Because sensitivity analysis can be computationally
expensive, we are developing methods that use automatic code differentiation to
efficiently compute the derivatives of variables with respect to each other,
and Bayesian optimization techniques to estimate sensitivity functions with as
few samples as possible. The final product of this analysis (a) includes
modular executable representations of grounded models (as dynamic Bayesian
networks), (b) provides results of model comparison to enable model choice in
tasks, and (c) based on grounded model similarity and differences, enables
model composition.

\begin{figure}
  \includegraphics[width=\textwidth]{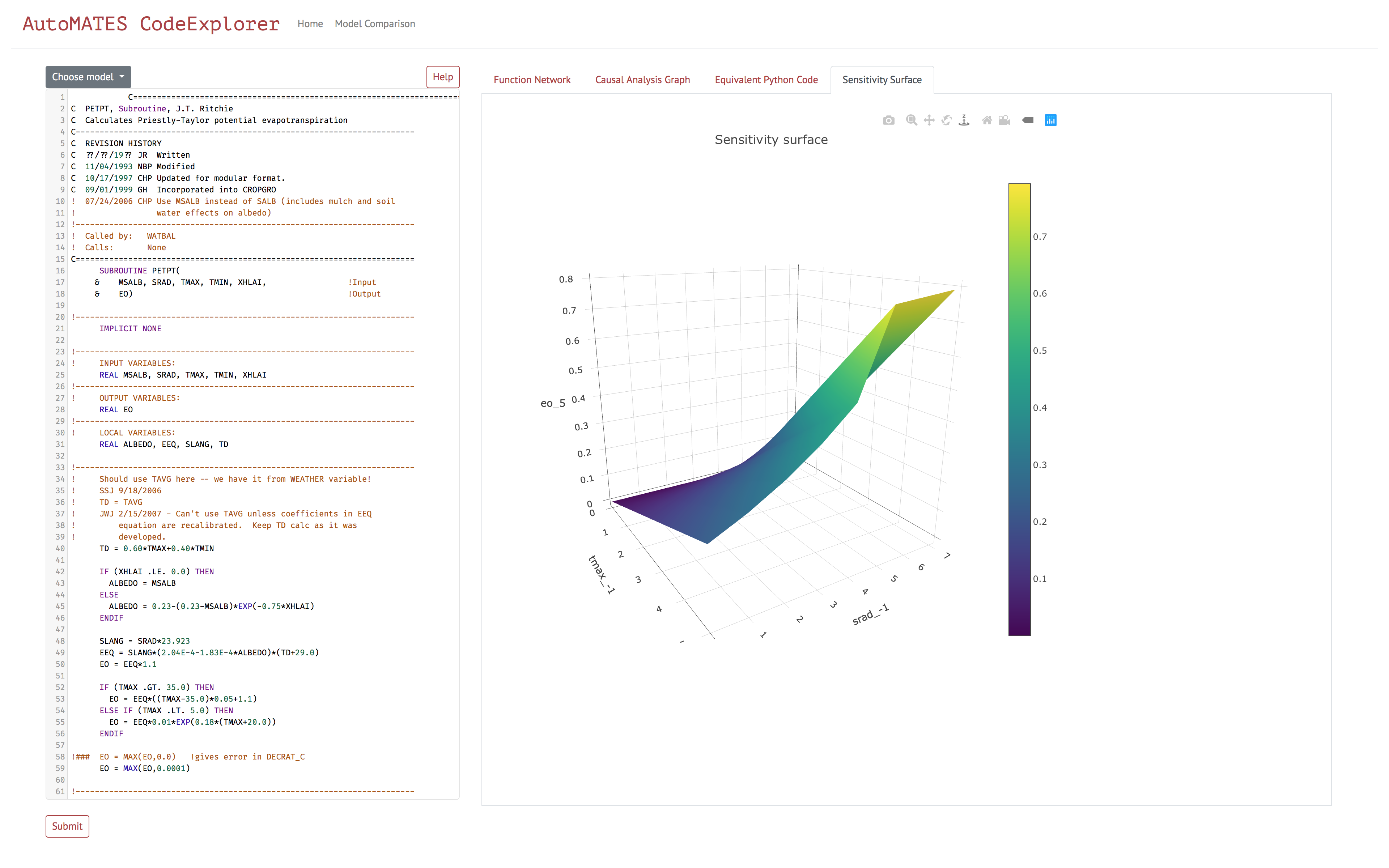}
  \caption{Initial results of automated sensitivity analysis. The pair of
    variables that the Priestley-Taylor model of evapotranspiration is most
    sensitive to has been automatically identified given bounds information
    for the input variables, and a surface plot has been generated that shows
  the effect of varying that pair of variables (maximum temperature and solar
radiation) on the output variable (potential evapotranspiration).}
  \label{fig:sensitivity_surface}
\end{figure}

In \autoref{fig:sensitivity_surface}, we show some initial results from
automated sensitivity analysis.

\section{Conclusions}
\label{sec:conclusion}

Systems of interest for scientific, humanitarian, and security reasons often
require the integration of computational models from multiple domains - for
example, modeling food security in a region requires the use of computational
crop, weather, and hydrology models, to name but a few. However,  this
integration currently requires significant manual effort in the form of
exposing and curating interfaces to the computational models. The framework we
are developing will greatly speed up this curation and integration process,
making it possible to effectively model large, complicated systems and reason
about them at multiple levels of abstraction.

\section{Resources, web sites, etc.}
\label{sec:resources}

The system described here is open-source and publicly available at
\href{https://github.com/ml4ai/automates}{\texttt{github.com/ml4ai/automates}} and
\href{https://github.com/ml4ai/delphi}{\texttt{github.com/ml4ai/delphi}}. We have also
set up a public webapp, CodeExplorer (see screenshot in
\autoref{fig:demowebapp}), which shows off a subset of the functionality of the
AutoMATES system, and is live at
\href{http://vanga.sista.arizona.edu/automates}{\texttt{vanga.sista.arizona.edu/automates}}.

\section{Acknowledgments}
This work is supported by the Defense Advanced Research Projects Agency (DARPA)
as part of the Automated Scientific Knowledge Extraction (ASKE) program under
agreement number HR00111990011.

\bibliography{bibliography}

\begin{thebibliography}{10}

\bibitem{bruce2014mathematical}
Jacob~Robert Bruce.
\newblock Mathematical expression detection and segmentation in document
  images.
\newblock Master's thesis, Virginia Tech, 2014.

\bibitem{Camargo:2016}
G.~G.~T. Camargo and A.~R. Kemanian.
\newblock Six crop models differ in their simulation of water uptake.
\newblock {\em Agricultural and Forest Meterology}, 220:116--129, 2016.

\bibitem{chu2013mathematical}
Wei-Ta Chu and Fan Liu.
\newblock Mathematical formula detection in heterogeneous document images.
\newblock In {\em 2013 Conference on Technologies and Applications of
  Artificial Intelligence (TAAI)}, pages 140--145, 2013.

\bibitem{pdffigures2}
Christopher Clark and Santosh Divvala.
\newblock Pdffigures 2.0: Mining figures from research papers.
\newblock 2016.

\bibitem{Deng:2017}
Yuntian Deng, Anssi Kanervisto, Jeffrey Ling, and Alexander~M. Rush.
\newblock Image-to-markup generation with coarse-to-fine attention.
\newblock In {\em Proceedings of the 34th International Conference on Machine
  Learning}, pages 980--989, 2017.

\bibitem{Deng:2016}
Yuntian Deng, Anssi Kanervisto, and Alexander~M. Rush.
\newblock What you get is what you see: {A} visual markup decompiler.
\newblock {\em CoRR}, abs/1609.04938, 2016.

\bibitem{DSSAT}
J.W Jones, G~Hoogenboom, C.H Porter, K.J Boote, W.D Batchelor, L.A Hunt, P.W
  Wilkens, U~Singh, A.J Gijsman, and J.T Ritchie.
\newblock The dssat cropping system model.
\newblock {\em European Journal of Agronomy}, 18(3):235 -- 265, 2003.
\newblock Modelling Cropping Systems: Science, Software and Applications.

\bibitem{redmon2016you}
Joseph Redmon, Santosh Divvala, Ross Girshick, and Ali Farhadi.
\newblock You only look once: Unified, real-time object detection.
\newblock In {\em Proceedings of the IEEE conference on computer vision and
  pattern recognition}, pages 779--788, 2016.

\bibitem{Escarcega:2018}
Marco~A. Valenzuela-Esc{\'a}rcega, {\"O}zg{\"u}n Babur, Gus Hahn-Powell, Dane
  Bell, Thomas Hicks, Enrique Noriega-Atala, Xia Wang, Mihai Surdeanu, Emek
  Demir, and Clayton~T. Morrison.
\newblock Large-scale automated machine reading discovers new cancer driving
  mechanisms.
\newblock {\em Database: The Journal of Biological Databases and Curation},
  2018.

\bibitem{valenzuela2016odin}
Marco~A. Valenzuela-Esc{\'a}rcega, Gus Hahn-Powell, and Mihai Surdeanu.
\newblock Odin's runes: A rule language for information extraction.
\newblock In {\em 10th International Conference on Language Resources and
  Evaluation}, 2016.

\end{thebibliography}
\bibliographystyle{plain}

\end{document}